\documentclass[10pt,twocolumn,letterpaper]{article}

\usepackage[pagenumbers]{cvpr} % To force page numbers, e.g. for an arXiv version

\usepackage{amsmath}
\usepackage{mathtools}
\usepackage{amsthm}
\usepackage{amssymb}
\usepackage[table]{xcolor}
\usepackage{colortbl}

\usepackage{enumitem}
\usepackage{multirow}
\usepackage{float}
\usepackage{bm}

\def\eg{\emph{e.g.}}

\def\ie{\emph{i.e.}}
\definecolor{hl}{RGB}{240,240,240}
\definecolor{comment}{RGB}{100,100,100}

\usepackage[utf8]{inputenc}
\DeclareUnicodeCharacter{2212}{-}
\definecolor{cvprblue}{rgb}{0.21,0.49,0.74}
\usepackage[pagebackref,breaklinks,colorlinks,allcolors=cvprblue]{hyperref}

\def\confName{CVPR}
\def\confYear{2026}

\title{Off-the-shelf Vision Models Benefit Image Manipulation Localization}

\author{
Zhengxuan Zhang$^1$ \quad Keji Song$^1$ \quad Junmin Hu$^1$ \quad Ao Luo$^2$ \quad Yuezun Li$^{1,}$\thanks{Corresponds to Yuezun Li (\url{liyuezun@ouc.edu.cn})} \;   \\
$^1$ School of Computer Science and Technology, Ocean University of China \\
$^2$  Southwest Jiaotong University 
}

\begin{document}
\maketitle

\begin{abstract}
   Image manipulation localization (IML) and general vision tasks are typically treated as two separate research directions due to the fundamental differences between manipulation-specific and semantic features. In this paper, however, we bridge this gap by introducing a fresh perspective: these two directions are intrinsically connected, and general semantic priors can benefit IML.    
   Building on this insight, we propose a novel trainable adapter (named \textbf{ReVi}) that \underline{re}purposes existing off-the-shelf general-purpose \underline{vi}sion models (\eg, image generation and segmentation networks) for IML. 
   Inspired by robust principal component analysis, the adapter disentangles semantic redundancy from manipulation-specific information embedded in these models and selectively enhances the latter. Unlike existing IML methods that require extensive model redesign and full retraining, our method relies on the off-the-shelf vision models with frozen parameters and only fine-tunes the proposed adapter. The experimental results demonstrate the superiority of our method, showing the potential for scalable IML frameworks. Codes are available at: \url{https://github.com/OUC-VAS/ReVi}
\end{abstract}    
\section{Introduction}
\label{sec:intro}

Digital image manipulation typically involves operations such as copy-move, splicing, and object removal.\footnote{Copy-move refers to cropping a region from an image and pasting it into a different location within the same image, whereas splicing involves inserting a region taken from a different image. Object removal means removing a target object and then inpainting the removed area with a visually harmonious background.} With the rapid advancement of image editing tools (\eg, Photoshop) and AI generative techniques (\eg, {GAN and diffusion models}), manipulated content has become significantly realistic and can easily bypass human visual inspection. The misuse of such manipulations can introduce widespread misinformation and fabricated content, posing serious threats to cybersecurity~\cite{imc1,imc2,imc3,imc4}. Consequently, image manipulation localization (IML) has emerged as an urgent and critical task. 

\begin{figure}[!t]
    \centering
    \includegraphics[width=0.85\linewidth]{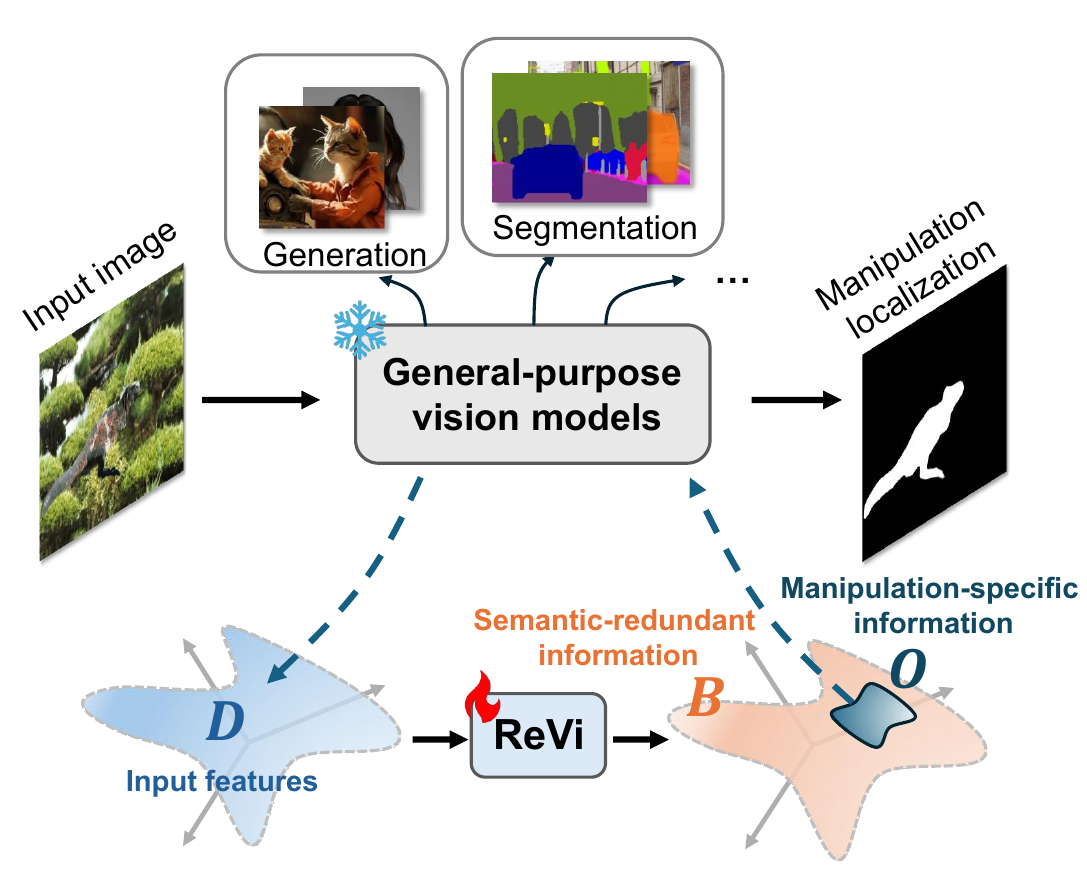}
    \vspace{-0.3cm}
    \caption{\small Overview of the proposed method (ReVi). The adapter can be integrated into the off-the-shelf general-purpose vision models and repurposes them for IML by decomposing semantic-redundant and manipulation-specific information.}
    \label{fig:overview}
    \vspace{-0.4cm}
\end{figure}

IML refers to identifying whether each pixel in an image is authentic or manipulated. From this definition, it is easy for beginners to mistakenly view IML as a special case of semantic segmentation. However, these two tasks differ fundamentally, as manipulation traces are significantly more subtle than explicit semantic cues. Consequently, IML is typically treated as an independent research direction, where specialized architectures are designed and trained on dedicated public benchmarks. With continuous effort, these methods have shown encouraging performance~\cite{SPAN,imlvit,mesorch}.

\begin{figure*}
    \centering
    \includegraphics[width=\linewidth]{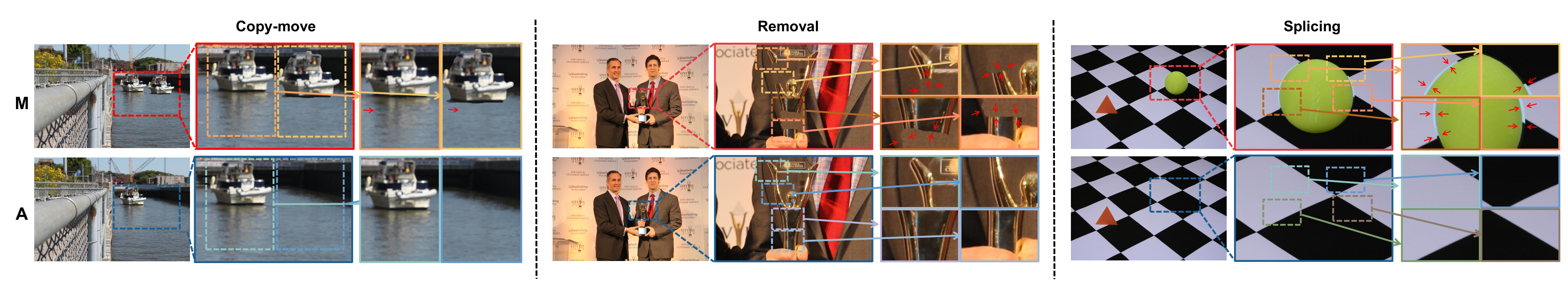}
    \vspace{-1.2cm}
    \caption{\small Top and bottom rows denote manipulated and authentic images. It can be seen that manipulations often cause semantic inconsistencies between the forged and the surrounding area. For example, copy-move object lacks its reflection on the water surface (left), removal makes the trophy discontinuous (middle), and splicing disrupts the background continuity around the sphere (right). }
    \label{fig:bond}
\end{figure*}

In this paper, we move beyond this conventional paradigm by raising a fresh hypothesis: \textit{semantic information and manipulation traces are likely connected, and semantic priors benefit IML}. This intuition can be preliminarily illustrated by examples in Fig~\ref{fig:bond}, where manipulations often introduce semantic inconsistencies. For instance, For example, copy-move object lacks its reflection on the water surface (Fig~\ref{fig:bond} left), removal makes the trophy discontinuous (Fig~\ref{fig:bond} middle), and splicing disrupts the background continuity around the sphere (Fig~\ref{fig:bond} right). Given that the research on general-purpose vision models has matured significantly, resulting in models that are both well studied and highly capable, a natural question arises: \textit{Can off-the-shelf vision models be effectively adapted for IML?} 
Investigating this question is particularly meaningful, as effectively leveraging such models for IML would remove the need to design and fully retrain task-specific architectures, thereby enabling a scalable and resource-efficient solution.

% Our study is motivated by one key observation and one central hypothesis. The observation is that the research community surrounding general-purpose vision models has become large and mature, resulting in models that are both well studied and highly capable. The hypothesis is that, although general semantic knowledge is not explicitly related to IML, it can nonetheless facilitate contextual understanding of images and provide implicit cues for manipulation localization, as manipulated objects may not be semantically matched with the surroundings. 

To achieve this goal, we propose a novel trainable adapter (named \textbf{ReVi}) that \underline{re}purposes existing off-the-shelf general-purpose \underline{vi}sion models (\eg, image generation and segmentation networks) for IML. \textit{Unlike typical adaptation strategies that primarily capture downstream semantic knowledge~\cite{sem1,sem2}, designing ReVi is more challenging, as it must extract and learn subtle manipulation traces under the guidance of semantic priors}. Specifically, ReVi is inspired by robust principal component analysis (RPCA), reformulated for IML by decomposing semantic-redundant and manipulation-specific information, and subsequently enhances vision models with the latter. To enable seamless integration, this formulation is implemented by deep neural operations and consists of two modules: Semantic Knowledge Approximation (SKA) and Manipulation Knowledge Enhancement (MKE). SKA is designed to model the dominant and authentic semantic knowledge, analogous to the principal components in PCA. In contrast, MKE focuses on isolating sparse and anomalous features that potentially correspond to manipulated regions, akin to the residual components. By operating cooperatively, these modules capture and amplify manipulation-specific information, effectively enhancing IML performance.

Our method is validated on various vision models, including image generation and segmentation models. 
Experimental results demonstrate that our method works as a plug-and-play module and enables effective adaptation of vision models for IML, offering a new insight into the following research. In summary, our contributions are threefold:
\begin{itemize}
    \item We introduce and demonstrate an interesting hypothesis that general semantic knowledge can benefit image manipulation localization, bridging the gap between IML and off-the-shelf vision models.
    \item We propose a novel adapter (\textbf{ReVi}) that reformulates principal component analysis for IML to extract manipulation-specific knowledge from vision models.
    \item Extensive experiments on various vision models corroborate the efficacy of the proposed adapter, showing promising performance compared to mainstream IML methods.
\end{itemize}

\section{Related Work}
\label{sec:formatting}

\smallskip\noindent\textbf{Image Manipulation Localization} (IML) is a long-standing problem in digital forensics that has constantly attracted attention over the past decade. Driven by the rapid advances in image editing techniques, IML has become increasingly important for societal security. Unlike traditional vision tasks such as semantic segmentation, which focus on explicit semantic content, IML targets subtle manipulation traces, including edge artifacts, pattern inconsistencies, and texture discontinuities. As a result, IML has evolved into an independent research direction.

Most existing IML methods design task-specific deep neural network architectures, \eg,~\cite{SPAN,objectformer,imlvit,mesorch, sparsevit}. These methods typically adopt an encoder–decoder framework, where the encoder extracts manipulation traces from the image, and the decoder generates a pixel-level localization map of the manipulated regions. To enhance the performance, a variety of auxiliary modules are proposed to leverage different cues, including JPEG compression artifacts~\cite{catnetv2,DiRLoc}, frequency statistics~\cite{HIFInet,objectformer}, noise inconsistencies~\cite{noiseguidance,trufor}, multi-scale spatial features~\cite{pscc}, contrastive learning~\cite{ACBG,exploring}, or specialized attention mechanisms~\cite{adaifl}. We note that Mesorch~\cite{mesorch} also emphasizes the importance of semantic consistency for IML. However, this method treats it only as an auxiliary cue and still relies on designing and fully training task-specific architectures to capture both semantic and manipulation features.

Despite their effectiveness, the aforementioned methods generally require end-to-end training on IML datasets, which introduces substantial computational costs. In contrast, several studies explore the use of large pretrained vision models, such as SAM~\cite{SAM}, by introducing additional modules or partially fine-tuning the backbone~\cite{ForensicsSAM, 2025Noise, detectiveSAM}. SAFIRE~\cite{safire} fully fine-tunes SAM  using six RTX 4090 GPUs, which limits practicality for users. In addition to leveraging large vision models, FakeShield~\cite{fakeshield} also incorporates multimodal large language models to provide explanatory support for image localization. EVP~\cite{evp} is designed for foreground segmentation by adding a number of adapters on pretrained ViTs, including SAM. Besides tuning adapters, this method also needs to fully fine-tune the decoder. While it can be extended to IML, its performance remains suboptimal.

% \smallskip
\smallskip\noindent\textbf{Robust Principal Component Analysis} (RPCA) is an extension of classical Principal Component Analysis designed to handle data corrupted by outliers~\cite{robpca,rpca-lc}. Unlike standard PCA, RPCA explicitly models observations as the superposition of two components: a low-rank component that captures the globally consistent structure, and a sparse component that accounts for anomalous patterns. Owing to this property, RPCA can be adopted in many vision tasks, such as image processing, classification, segmentation, object detection, etc.~\cite {10.1137/100817206,6819824,8017459,7464858}. 

However, solving RPCA is non-trivial, which motivates a line of research that leverages deep neural networks to approximate RPCA solutions efficiently~\cite{gregor2010learning,zhang2018ista,yan2023multispectral,Wu_2024_WACV,wu2025rpcanet_pp}. Nevertheless, RPCA has not yet been extended to other domains such as IML. Based on our observations, the decomposition property of RPCA is particularly well-suited for IML, as extracting subtle manipulation traces from clean images can be viewed as a special case of RPCA. Therefore, in this paper, we take a fresh step by reformulating RPCA in an appendable manner for general-purpose vision models and repurposing them as effective image manipulation localizers.

\section{Method}

\subsection{Hypothesis and Preliminary Investigation}
We hypothesize that the general semantic knowledge can benefit the IML task. It is motivated by the observation that image manipulations often disrupt the coherent semantic context of an image. As illustrated in Fig.~\ref{fig:bond}, spliced objects usually introduce artifacts across their boundaries, breaking semantic continuity. Although general semantic knowledge may not explicitly encode such abnormalities, it implicitly encapsulates these inconsistencies, thereby facilitating the exposure of manipulated regions.

To validate this hypothesis, we conduct preliminary experiments using Latent Diffusion Model (LDM)~\cite{ldm}, a well-established model for general-purpose image generation. And TinySAM, a powerful model for semantic segmentation. The model we use is originally trained on a subset of the LAION-5B dataset~\cite{laionb} (containing images of multiple categories such as persons, objects, and scenes) and then partially fine-tuned for the IML task using Low-Rank Adaptation (LoRA)~\cite{lora}, a classical parameter-efficient fine-tuning (PEFT) strategy widely adopted for large-scale models.

\smallskip\noindent\textbf{LoRA.} Instead of updating the entire weight matrix $W\in \mathbb{R}^{d \times k}$ of a pre-trained layer, LoRA constrains the weight update by a low-rank decomposition. It injects trainable rank-decomposition matrices into the model architecture. For a given pre-trained weight matrix $W_{0}$, the update process can be written as
\begin{equation}
    W = W_0 + \Delta W  = W_0  + B A ,
\end{equation}
where $A\in \mathbb{R}^{r \times k}$ and $B\in \mathbb{R}^{d \times r}$ are the trainable low-rank matrices, and the rank $r\ll \min (d,k)$. During training, $W_{0}$ is frozen (\ie, its gradients are not computed), and only $A$ and $B$ are updated, dramatically reducing the number of trainable parameters. Recent works typically incorporate this strategy into the attention mechanisms of Transformer blocks~\cite{ComputationalloraLimits,advancing}.

\smallskip\noindent\textbf{Preliminary Experiments.} 
% The LDM model consists of an auto-encoder and a Transformer-based U-Net. \red{For unconditional input, }the \red{auto-}encoder maps the input image into latent representations, on which the U-Net performs the diffusion process. Then the \red{auto-}decoder reverts the latent representations back to image space. 
% In our experiment, all original parameters are frozen, and LoRA is modules are inserted into \red{all convolutional and linear layers} of U-Net \red{and auto-encoder} for parameter-efficient fine-tuning. 
In our experiment, all original parameters are frozen, LoRA is used on the U-Net and auto-encoder structure of LDM for parameter-efficient fine-tuning. 
The resulting model is fine-tuned on the PSCC dataset~\cite{pscc}, a widely used pretraining dataset for IML. For comparison, we additionally conduct a baseline experiment in which the entire model is fully fine-tuned on the PSCC dataset with all parameters unfrozen. Both models are then evaluated on the CASIAv1 dataset~\cite{casia}, a commonly adopted protocol for IML evaluation. 
The results are reported in Table~\ref{tab:preliminary1}. The \textbf{top} part denotes the dedicated IML methods, while the \textbf{middle} part corresponds to LDM with full fine-tuning, LoRA, and our method ReVi. It can be observed that 
\begin{itemize}
    \item Using LoRA can achieve decent performance compared to IML dedicated methods, highlighting the potential of enabling a general vision model for IML. By adding our method ReVi, the performance is further improved. 
    \item Fully fine-tuning the LDM does not outperform the LoRA strategy. We conjecture that compared to LoRA, full fine-tuning could largely disrupt its original semantic knowledge, thereby resulting in a performance drop. 
    % \item \red{While LoRA still has room for improvement in model performance on IML tasks, our designed ReVi module can better collaborate with LoRA to further enhance localization accuracy.}
\end{itemize}

Moreover, to demonstrate the difference between the IML and general vision tasks,  we select a powerful semantic segmentation model, TinySAM, and conduct two experiments: First, training it on the IML dataset from scratch without using its pre-trained weights. Second, freezing its pre-trained weights and fine-tuning it using LoRA. The results reveal that 
\begin{itemize}
    \item Without the pretrained weights, TinySAM does not perform well, highlighting the gap between general vision tasks and IML. 
    \item When using its pretrained weights unchanged, performance improves largely, demonstrating the importance of strong semantic priors for IML.
\end{itemize}
These findings align with our hypothesis that general semantic knowledge is beneficial for the IML task.

% To further demonstrate the difference between the IML task and traditional vision tasks as noted in the introduction, we selected the TinySAM model framework—widely used in vision tasks such as segmentation—and trained it on the IML task directly without loading its pre-trained weights. We compared this with LoRA fine-tuning using the pre-trained weights. Both models were trained on the PSCC dataset and tested on the CASIAv1, Columbia, Coverage, NIST16, and IMD20 datasets. The results in Table~\ref{tab:preliminary2} show that, except for a slight improvement on the NIST16 dataset by the former, the performance of direct training lags significantly behind that of fine-tuning with the preserved original model weights on the vast majority of datasets. This demonstrates that frameworks suitable for semantic vision tasks are not directly applicable to the IML task, and that preserving the original semantic information is beneficial for IML.
% These findings align with our hypothesis that general semantic knowledge is beneficial for the IML task.

\begin{table}[!t]
    \centering
    \caption{\small Preliminary investigation on IML task using Latent Diffusion Model (LDM) and TinySAM. Following the protocols in previous works~\cite{pscc,mgq}, these methods are pretrained on the PSCC dataset and directly evaluated on the CASIAv1 dataset with F1 score. ``Full FT'' denotes full fine-tuning. ``LoRA'' and ``ReVi'' represent partial fine-tuning using the respective adapters.}
    \vspace{-0.2cm}
    \begin{tabular}{ll}
    \toprule
         \multirow{1}{*}{Method} & CASIAv1 \\
         \midrule
         PSCC (CVPR'22)~\cite{pscc}  & 28.83\\ 
         EVP (CVPR'23)~\cite{evp}  & {36.56}\\
         Mesorch (AAAI'25)~\cite{mesorch} & 32.09 \\
         \midrule
         LDM \textit{w.}  Full FT & 29.75 \\ 
         LDM \textit{w.}  LoRA & 33.17  \textcolor{violet}{(+ 3.42)} \\
         LDM \textit{w.} \textbf{ReVi}   & 39.61  \textcolor{violet}{(+ 9.86)} \\
         \midrule
         TinySAM \textit{wo.} pretrained weights    & 22.87 \\
         TinySAM \textit{w.} pretrained weights \textit{w.} LoRA  & 33.63 \textcolor{violet}{(+ 10.76)} \\
        %  \midrule
        % \red{LDM \textit{w.} ReVi} & 39.61 & \textcolor{violet}{(+ 6.--)} \\
         \bottomrule
    \end{tabular}
    \label{tab:preliminary1}
\end{table}

\subsection{The Proposed Adapter}
% Motivated by the preliminary investigation, we propose a RPCA-inspired adapter that leverages the general semantic knowledge while further captures manipulation-specific information. 

\smallskip\noindent\textbf{Revisit of RPCA.}
Robust Principal Component Analysis (RPCA) is extended from the traditional PCA algorithm, which is more effective in representing outliers. Specifically, RPCA decomposes the image matrix into a low-rank matrix with background information and a sparse matrix with object-related information. Owing to this property, RPCA can be adopted into many vision tasks, including image classification, foreground segmentation, and object detection~\cite{5995484,8017459,7464858}.
Denote the input image as $\bm{D}$, the background information as $\bm{B}$, and the object-related information as $\bm{O}$. The RPCA decomposition can be expressed as $\bm{D} = \bm{B} + \bm{O}$. For optimization, an iterative framework with respect to $\bm{B}$ and $\bm{O}$ is introduced~\cite{Wu_2024_WACV,wu2025rpcanet_pp} (see Fig.~\ref{fig:RPCA}), where $\mathrm{prox}$ denotes a proximal operator, ${\rho}$ is a stage-independent learnable parameter and $\nabla \mathcal{T}$ a Lipschitz continuous gradient function. 

\begin{figure}[h]
    \centering
    \renewcommand\arraystretch{1.5}
    \begin{tabular}{l}
    \toprule
         $\mathrm{For} \; k = 1:K$  \\
         % \midrule
         % \hspace{0.5cm} \textcolor{comment}{// $\mathrm{prox}$ is a proximal operator} \\
         \hspace{0.5cm} $1. \; \bm{B}^k = \mathrm{prox}(\bm{D}^{k-1} - \bm{O}^{k-1});$\\
         % \hspace{0.5cm} \textcolor{comment}{// $C$ and $\epsilon$ are predefined constant} \\
         % \hspace{0.5cm} $\bm{W}^k = C / (|\bm{O}^{k-1}| + \epsilon);$ \\
         \hspace{0.5cm} $2. \; \bm{O}^k = \bm{O}^{k-1} + \bm{D}^{k-1} - \bm{B}^{k} - {\rho} \nabla \mathcal{T}(\bm{O}^{k-1});$ \\
         \hspace{0.5cm} $3. \; \bm{D}^k = \bm{B}^{k} + \bm{O}^k;$ \\
         \bottomrule
    \end{tabular}
    \label{fig:RPCA}
    \small
    \vspace{-0.2cm}
    \caption{\small Iterative optimization framework for RPCA.}
    \label{fig:RPCA}
\end{figure}

To efficiently solve this problem, several studies have proposed deep networks that approximate the iterative RPCA optimization~\cite{gregor2010learning,zhang2018ista,yan2023multispectral,Wu_2024_WACV,wu2025rpcanet_pp}. While they have demonstrated effectiveness on general vision tasks, their potential for IML has not yet been explored.

\begin{figure}[!t]
    \centering
    \includegraphics[width=\linewidth]{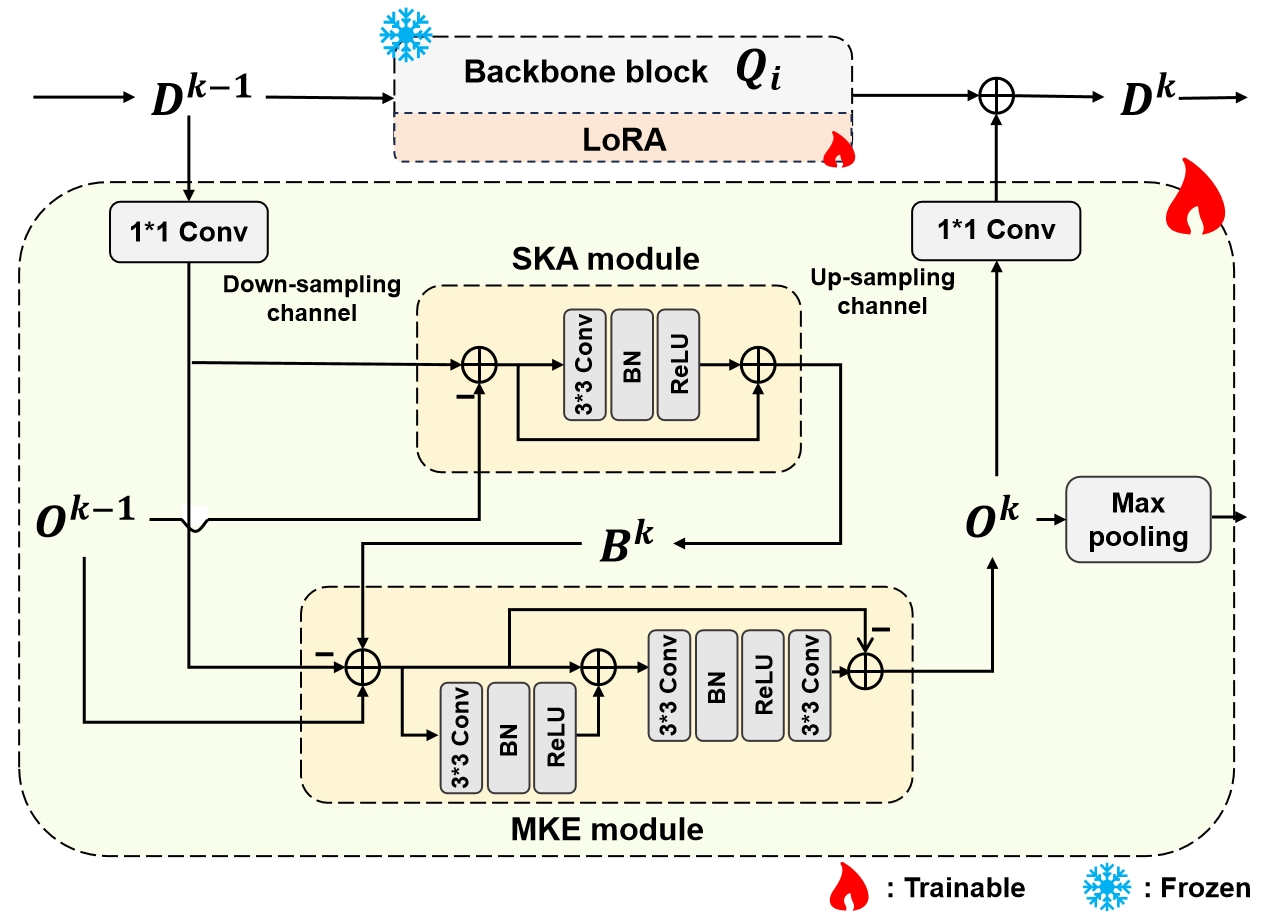}
    \vspace{-0.8cm}
    \caption{\small Overview of the proposed ReVi. The symbol ``$-$'' denotes the subtraction operation. See text for details.}
    \label{fig:method}
    % \vspace{-0.3cm}
\end{figure}
%The adapter The framework consists of three main components: the SKA module, which extracts background information; the MKE module, which captures tampered target information; and the backbone network $Q_i$. Here, 

\smallskip\noindent\textbf{Adapter Design.} 
We design a specific deep learning architecture that satisfies the following two criteria: 
\begin{enumerate}
    \item The adapter is dedicated for IML task. Following the principle of RPCA, we reformulate the features extracted from a backbone model (\eg, LDM) as $\bm{D}$, and expect to decompose it into semantic-redundant information $\bm{B}$ and manipulation-specific information $\bm{O}$. The latter is then leveraged to enhance general-purpose vision models.
    \item The adapter can be elegantly integrated into general-purpose vision models, improving the performance while introducing only modest computational overhead.
\end{enumerate}
Specifically, our adapter contains two modules: Semantic Knowledge Approximation (SKA) and Manipulation Knowledge Enhancement (MKE), which are designed to approximate semantic-redundant knowledge and extract manipulation-specific information. The overview of this adapter is illustrated in Fig.~\ref{fig:method}. 

Denote backbone model as $\bm{F}$, which contains $n$ key blocks $\{\bm{Q}_i \}^n_{i=1}$. As shown, the adapter is appended on one block $\bm{Q}_i$ and the two modules work cooperatively. At iteration $k$, the SKA module first infers the current background representation $\bm{B}^k$ from the input features $\bm{D}^{k-1}$ provided by the backbone model and the manipulation features $\bm{O}^{k-1}$ from the previous step. It then passes all three components to the MKE module, which subsequently isolates the updated manipulation features $\bm{O}^k$ by training with task-specific objectives. The features $\bm{O}^k$ are then used to amplify the manipulation traces within the backbone model features. This RPCA-inspired adapter is driven by the iterative update from $\bm{D}^{k-1}$, $\bm{O}^{k-1}$ to $\bm{O}^k$, enhancing the manipulation traces within general semantic features, allowing for effective IML without fine-tuning the original model.

\smallskip\noindent\textbf{Semantic Knowledge Approximation.} 
This module aims to predict the semantic-redundant content from the current features. It implies the authentic content, whose semantic knowledge is largely consistent and thus provides limited cues for exposing manipulations. 
Existing works such as ISTANet++~\cite{you2021ista} and RPCANet++~\cite{wu2025rpcanet_pp} leverage complex architectures (\eg, ConvLSTM) to simulate the $\mathrm{prox}$ operations in equation (1) of Fig.~\ref{fig:RPCA}. Considering the efficiency, in our case, we only use a single convolutional layer with kernel size $3 \times 3$ and a skip connection for simulation, which can be formulated as 
\begin{equation}
\renewcommand{\arraystretch}{1.2}
\begin{array}{rl}
   \bm{B}^{k}  &  = \Psi(\bm{D}^{k-1},\bm{O}^{k-1}) \\
     & = \mathrm{Conv}(\bm{D}^{k-1}-\bm{O}^{k-1})+\bm{D}^{k-1}-\bm{O}^{k-1}.
\end{array}
\end{equation}

% \begin{equation}
%   \bm{B}^{k} = \mathrm{ReLU}\left(\mathrm{BN}\left(\mathrm{Conv}_{3\times3}(D^{k-1}-O^{k-1})\right)\right)+D^{k-1}-O^{k-1},
% \end{equation}

% While RPCANet++ employs a structure analogous to ConvLSTM, its experiments indicate that a purely convolutional residual with residual construction architecture yields comparable performance. Hence, we discard this complex mechanism and instead adopt a simple convolutional layer to simulate the process, thereby further reducing computational overhead. The process of SKA can be formulated as:
% \begin{equation}
%   B^{k} = \mathrm{ReLU}\left(\mathrm{BN}\left(\mathrm{Conv}_{3\times3}(D^{k-1}-O^{k-1})\right)\right)+D^{k-1}-O^{k-1},
% \end{equation}
% where $k$ indicates the position of the PiA module. This formulation closely corresponds to the derivation of \eqref{b*result} and retains the residual structure, thereby enhancing stability. 

\smallskip\noindent\textbf{Manipulation Knowledge Enhancement.} 
% The MKE module is designed to separate potential manipulation information from the feature representations. As described in \eqref{o*result}, by setting $\lambda$ to 0.5, we obtain the following expression:
% \begin{equation}
%   O^{k} = O^{k-1} + D^{k-1} - B^{k} - \rho^{k} \nabla S^{k}(O^{k-1})\space.
% \end{equation}
Given the equation (2) of Fig.~\ref{fig:RPCA} and $\bm{O}^{k−1}$, we preserve the difference between the previous $\bm{D}^{k−1}$ and the current approximation $\bm{B}^{k}$, and obtain the input feature as $\bm{X}^{k}=\bm{O}^{k−1}+\bm{D}^{k−1}−\bm{B}^{k}$ for MKE. Then we construct a two-layer convolutional network $\Phi$ to simulate the operator $\nabla \mathcal{T}$. To enhance the effect of $\bm{O}^{k−1}$ in $\nabla \mathcal{T}$, we use the difference between last $\bm{D}^{k-1}$ and current $\bm{B}^k$ with a learnable deviation $\bm{\Delta}$. Considering that $\Delta$ should be adaptive to the features, we use a single convolutional layer with a kernel of $3 \times 3$ for simulation. It is important to note that, supervised by task-specific objectives, $\Delta$ could extract manipulation-specific information. The overall formulation can be defined as
\begin{equation}
\renewcommand{\arraystretch}{1.2}
\begin{array}{rl}
     \bm{O}^k & = \bm{X}^{k} - {\rho} \nabla \mathcal{T}(\bm{O}^{k−1}+\bm{D}^{k−1}−\bm{B}^{k} + \bm{\Delta}) \\
     & = \bm{X}^{k} - {\rho} \Phi (\bm{O}^{k−1}+\bm{D}^{k−1}−\bm{B}^{k} + \bm{\Delta}) \\
     & = \bm{X}^{k} - {\rho} \Phi (\bm{X}^{k} + \bm{\Delta}) \\
     & = \bm{X}^{k} - {\rho} \Phi (\bm{X}^{k} + \mathrm{Conv}(\bm{X}^{k})). \\
\end{array}
    \end{equation}
% \red{Where $\rho$ is a learnable coefficient.}

The output $\bm{O}^{k}$ is utilized in two ways. First, it is combined with the features from the backbone model for enhancement. Since in our task, the manipulation traces are subtle and less, we can approximate $\bm{B}^k$ using $\bm{D}^{k-1}$, \ie, $\bm{B}^k \approx \bm{Q}_i (\bm{D}^{k-1})$. In this way, the enhancement corresponds to equation (3) of Fig.~\ref{fig:RPCA}, as
\begin{equation}
    \bm{D}^{k} = \bm{Q}_i (\bm{D}^{k-1}) + \alpha \bm{O}^{k},
\end{equation}
where $\alpha$ is a learnable coefficient that adjusts the importance of $\bm{O}^{k}$.

Second, $\bm{O}^{k}$ is propagated as a hidden state to the next block $\bm{Q}_{i+1}$ that requires our adapter. 

In summary, these two modules function sequentially to isolate $\bm{B}$ and $\bm{O}$ from the feature of the backbone model. To make the backbone model better understand the enhanced features, we also employ LoRA on the backbone model. The detailed derivation of the framework is presented in the \textit{Supplementary}.
% The overview of our method is shown in Fig.~\ref{fig:placeholder} and more details are provided in \red{Supplementary}.

\smallskip\noindent\textbf{Objectives.}
To train our adapters, we retain the original training objectives of general-purpose vision models and directly apply them to the IML task. For image generation models such as LDM, we continue to use the original noise prediction objective and simply average the output across channels to produce a localization map when inference. For segmentation models such as SegFormer~\cite{SegFormer}, we modify the final layer of the segmentation head to output binary predictions, while still employing the original binary cross-entropy (BCE) loss.

Moreover, inspired by prior works~\cite{imlvit,cui2025forensics}, we employ an edge-aware loss to highlight manipulation boundaries. Specifically, the ground-truth manipulation boundary is obtained by applying dilation and erosion operations to the original ground-truth masks and computing their absolute difference. The predicted boundary is derived in the same manner, and a BCE loss is applied for supervision. The final training objective is formulated as
\begin{equation}
    \mathcal{L} = \mathcal{L}_{\mathrm{org}} + \lambda \cdot \mathcal{L}_{\mathrm{edge}},
\end{equation}
where $\lambda$ controls the relative contribution of the two loss terms.
\section{Experiments}
\begin{table*}[!t]
\centering
\small
\caption{\small F1 (\%) performance under Pre-trained and Fine-tuned protocols.}
\vspace{-0.4cm}
\resizebox{\textwidth}{!}{%
\begin{tabular}{l|cccccc|cccccc}
\toprule
\multirow{2}{*}{Methods} & \multicolumn{6}{c|}{\textbf{Pre-trained protocol}} & \multicolumn{6}{c}{\textbf{Fine-tuned protocol}} \\
\cmidrule(lr){2-7} \cmidrule(lr){8-13}
& CASIAv1 & Columbia & Coverage & NIST16 & IMD20 & Average & CASIAv1 & Columbia & Coverage & NIST16 & IMD20 & Average \\
\midrule
\rowcolor{hl} \multicolumn{13}{c}{\textit{Dedicated methods}} \\
    PSCC-Net (CVPR'22)~\cite{pscc}         & 28.83 & 12.23 & 13.35 & 21.20 & 16.49 & 18.42 & 56.20 & 94.38 & 48.55 & 64.14 & 43.78 & 61.41 \\
    TruFor (CVPR'23)~\cite{trufor}           & 23.49 & 38.76 & 14.91 & 21.06 & 17.10 & 23.06 & 40.80 & 88.03 & 59.04 & 63.85 & 42.24 & 58.79 \\
    EVP (CVPR'23)~\cite{evp}              & 36.56 & 29.06 & 26.67 & 23.75 & 20.44 & 27.30 & 50.00 & 93.89 & 50.97 & 77.60 & 45.79 & 63.65 \\
    IML-ViT (Arxiv'23)~\cite{imlvit}         & 28.19 & 60.61 & 19.86 & 24.76 & 12.95 & 29.27 & \textbf{66.28} & 96.80 & 58.14 & 80.65 & \textbf{53.21} & 71.02 \\
    ACBG (TIFS'24)~\cite{ACBG}             & 23.47 & 42.99 & 08.31 & 16.49 & 13.17 & 20.89 & 55.78 & 83.92 & 41.03 & 50.89 & 42.97 & 54.92 \\
    Sparse-ViT (AAAI'25)~\cite{sparsevit}       & \underline{42.57} & \underline{78.43} & 18.76 & 24.82 & 12.93 & 35.50 & 54.36 & 96.11 & \underline{69.74} & 75.99 & 51.84 & 69.61 \\
    MPC (TIFS'25)~\cite{mpc}              & 22.37 & 72.45 & 09.49 & 18.59 & 11.65 & 26.91 & 42.45 & 98.04 & 56.52 & 70.07 & 46.32 & 62.68 \\
    Mesorch (AAAI'25)~\cite{mesorch}          & 32.09 & 63.86 & 17.46 & 21.18 & 14.27 & 29.77 & 59.82 & 97.63 & 49.47 & 82.61 & 43.64 & 66.63 \\
\midrule
\rowcolor{hl} \multicolumn{13}{c}{\textit{Adapting general-purpose vision models}} \\
    TinySAM \textit{w.} \textbf{ReVi}   & 38.79 & 55.41 & 21.35 & 24.00 & 16.24 & 31.16 & 60.98 & 98.62 & \textbf{70.01} & \textbf{88.52} & \underline{52.76} & \textbf{74.18} \\
    LDM \textit{w.} \textbf{ReVi}       & 39.61 & \textbf{84.84} & 26.77 & 18.47 & 15.36 & \underline{37.01} & \underline{61.02} & 98.52 & 60.84 & \underline{86.87} & 51.77 & \underline{71.80} \\
    SegFormer \textit{w.} \textbf{ReVi} & \textbf{42.73} & 60.25 & \textbf{28.93} & \underline{28.73} & \textbf{26.20} & \textbf{37.37} & 56.48 & \underline{98.80} & 66.44 & 86.11 & 50.99 & 71.76 \\
    Swin Transformer \textit{w.} \textbf{ReVi} & 38.80 & 60.20 & \underline{27.50} & \textbf{30.74} & \underline{24.36} & 36.32 & 55.64 & \textbf{98.82} & 59.10 & 86.71 & 52.71 & 70.60 \\
\bottomrule
\end{tabular}%
}
\label{tab:f1-combined}
\end{table*}

\begin{figure*}[!t]
    \centering
    \includegraphics[width=0.95\linewidth]{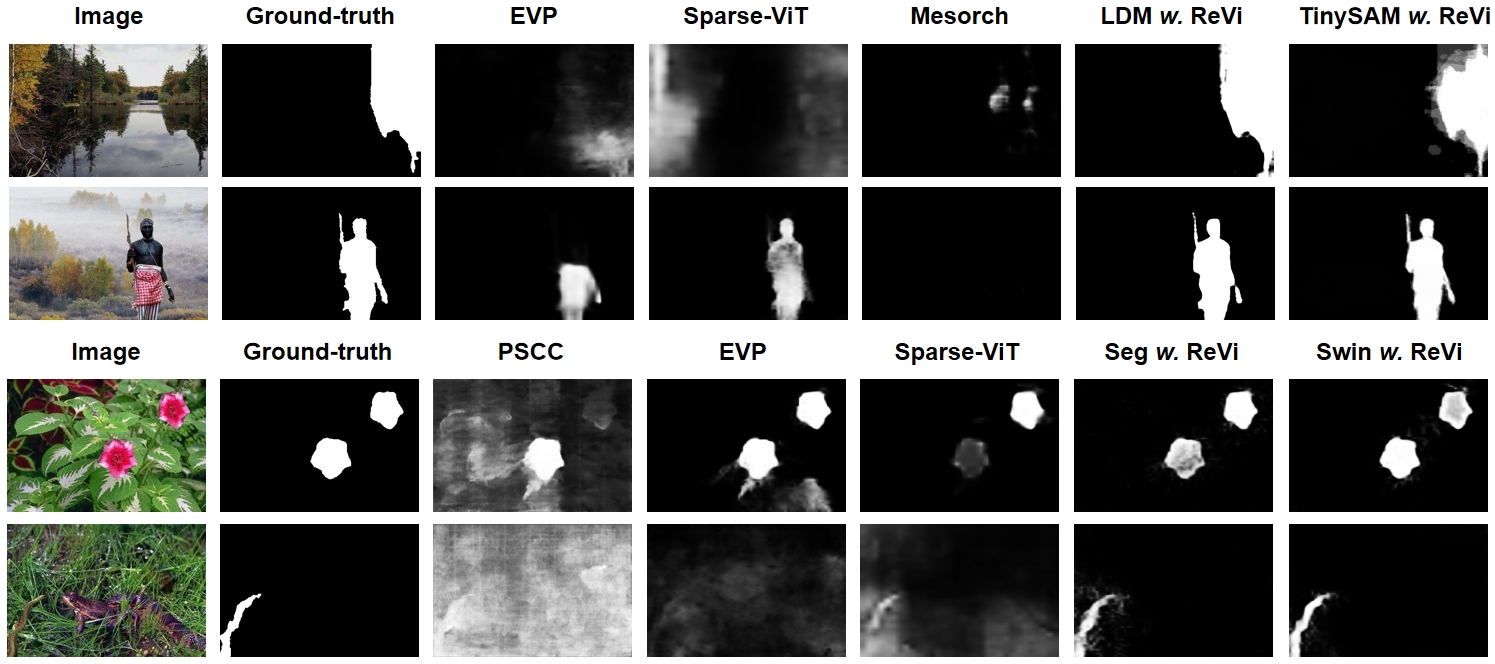}
    \vspace{-0.4cm}
    \small
    \caption{\small Visualization results comparing different methods on sample images. ``Seg'': SegFormer. ``Swin'': Swin Transformer.}
    \label{fig:vis-main}
    % \vspace{-0.3cm}
\end{figure*}

\subsection{Experimental Setup}

\smallskip\noindent\textbf{Datasets and Evaluation Metrics.} Our method is validated on five datasets: CASIA~\cite{casia}, Columbia~\cite{columbia}, Coverage~\cite{coverage}, NIST16~\cite{nist}, and IMD20~\cite{imd}. For CASIA, Coverage, and NIST16, we adopt the training–testing splits used in~\cite{pscc}. Since IMD20 and Columbia do not provide official splits, we randomly partition each dataset into training and testing subsets. Specifically, IMD20 is split in a 9:1 ratio, while the smaller Columbia dataset, which contains only 180 images, is split in a 7:3 ratio. More details of the datasets are shown in the \textit{Supplementary}.
Following previous works~\cite{imlvit, trufor, mvssnet,pscc,mgq}, our method is pretrained on the PSCC dataset~\cite{pscc}, and evaluated using the best F1 score. Our method is also evaluated using the AUC score. However, since the F1 score is considered more representative~\cite{imlvit, trufor, mvssnet}, the AUC results are provided in the \textit{Supplementary}.

\smallskip\noindent\textbf{Backbone Vision Models.} 
To demonstrate the effectiveness of our adapter, we choose four well-studied off-the-shelf pre-trained vision models: Latent Diffusion Model (LDM) ~\cite{ldm}, TinySAM~\cite{tinysam}, SegFormer~\cite{SegFormer}, and Swin Transformer~\cite{swin}, covering general image generation and segmentation tasks. These models are built upon Transformer-based architectures and have been pre-trained on large-scale natural image datasets, thereby acquiring comprehensive general semantic knowledge, which provides a solid foundation for the image manipulation localization task. Specifically, {LDM is pretrained on a subset of the LAION-5B dataset~\cite{laionb} for the text-to-image task. SegFormer is pretrained on the Pascal VOC dataset~\cite{voc12} for segmentation. Swin Transformer is pretrained on Image-1K dataset~\cite{imagenet1k} for segmentation. TinySAM is pretrained on $1\%$ of the SA-1B dataset~\cite{SAM} for segmentation. We emphasize that, due to limited computational resources, we adopt lightweight variants of each model: SegFormer-b1 and Swin-Tiny. Nevertheless, the results are surprisingly improved.

\smallskip\noindent\textbf{Implementation Details.} 
To reduce computational cost while improving deployment flexibility, all input images are resized to $256 \times 256$, a more practical setting in real-world scenarios. In training, we adopt the AdamW optimizer with an initial learning rate of $1 \times 10^{-4}$ and employ cosine annealing decay. The factor $\lambda$ in the objective is set to $20$. 
Specifically, for LDM, our adapter is added to the first self-attention module within each transformer block of the U-Net encoder. For TinySAM, it is integrated into the self-attention module of each transformer block in the image encoder. For SegFormer and Swin Transformer, the adapter is incorporated into the self-attention module of each transformer block in their encoders. All experiments are performed using the PyTorch framework on NVIDIA RTX 3090 GPUs. More implementation details are shown in the \textit{Supplementary}.
% \red{And the param rate are shown in Table~\ref{tab:param}}

\subsection{Comparison with Dedicated IML Methods}
While some recent methods demonstrate impressive performance, their results are often difficult to reproduce. For instance, certain works~\cite{HIFInet, noiseguidance, adaifl, 2024A, ForensicsSAM, detectiveSAM} have not released their training code, while others (e.g., SAFIRE\cite{safire}, FakeShield\cite{fakeshield}) require prohibitively extensive computational resources, utilizing six 4090 GPUs and four A100 GPUs, respectively. Additionally, to focus on comparing recent methods, some earlier works~\cite{catnetv2, objectformer, SPAN} are not included in this study. \textit{For a fair comparison, we select widely used baselines based on the following two principles: 1) the methods must have open-sourced complete training code, and 2) their required computational resources should be comparable to ours.} Accordingly, we include state-of-the-art methods from the past two years, such as ACBG~\cite{ACBG}, Sparse‑ViT~\cite{sparsevit}, MPC~\cite{mpc}, and Mesorch~\cite{mesorch}, as well as representative earlier works including PSCC-Net~\cite{pscc}, TruFor~\cite{trufor}, EVP~\cite{evp}, and IML‑ViT~\cite{imlvit}, as our comparative baselines. Then we retrain those methods under the same experimental settings. Deployment details of each method are provided in the \textit{Supplementary}.

Following previous works~\cite{pscc,mgq}, we evaluate all methods under two protocols: 1) Pre-trained protocol, where the methods are pretrained on the PSCC dataset and directly evaluated on other datasets, and 2) Fine-tuned protocol, where the pretrained methods are then fine-tuned on other datasets.

\smallskip\noindent\textbf{Pre-trained Protocol.} Table~\ref{tab:f1-combined} (left) shows the results of all methods under the Pre-trained protocol. 
% \blue{We observe that applying LoRA to vision models alone already achieves competitive performance compared with dedicated IML methods, which supports our hypothesis that general semantic knowledge benefits IML.} 
By adding our adapter (ReVi), the performance on SegFormer and LDM is further improved. Specifically, on SegFormer, ReVi achieves improvements of $1.9\%$ and $7.6\%$ in F1 score compared to the second-best method Sparse-ViT and the third-best method Mesorch, respectively; On LDM, the improvements are $1.5\%$ and $7.2\%$, respectively. These results demonstrate the good generalization capability of our method and its ability to capture manipulation traces.
%By adding our adapter (ReVi), the performance on SegFormer is further improved by $1.9\%$ and $7.6\%$ at F1 score, compared to the second-best method, Sparse-ViT, and the third-best method, Mesorch, demonstrating the good generalization capability and the ability of our method in capturing manipulation traces.
%the performance {on SegFormer} is further improved by $5.1\%$ at AUC and $1.9\%$ at F1 score, compared to the second-best method Sparse-ViT

\smallskip\noindent\textbf{Fine-tuned Protocol.} The results in Table~\ref{tab:f1-combined} (right) exhibit a trend similar to that observed in Pre-trained protocol, indicating that vision models with minimal adaptation can be effective for image manipulation localization. 
After fine-tuning on the respective datasets, our method consistently outperforms dedicated IML methods and achieves performance on par with, or surpassing, the second-best method, IML-ViT. These results further highlight the potential of our method for scalable IML. 
We observe that ReVi achieves the best performance in nearly all scenarios, except for the case when compared to IML-ViT on the two larger datasets, CASIAv1 and IMD20. We attribute this to the fact that under the fine-tuning protocol, the training and test sets exhibit similar distributions. Hence, even without the assistance of semantic knowledge, models can still overfit the test distribution effectively as long as sufficient data is available.

\smallskip\noindent\textbf{Visual Examples.} {Fig.~\ref{fig:vis-main} presents several visualization examples of different methods on CASIAv1. The results indicate that our method achieves superior localization accuracy compared to other methods, particularly in boundary regions.} {Fig.~\ref{fig:cam} displays the GradCAM of TinySAM on images, revealing that our ReVi guides the model to focus on manipulated regions. All examples are obtained under Fine-tuned protocol.}

\begin{figure}[!t]
    \centering
    \includegraphics[width=\linewidth]{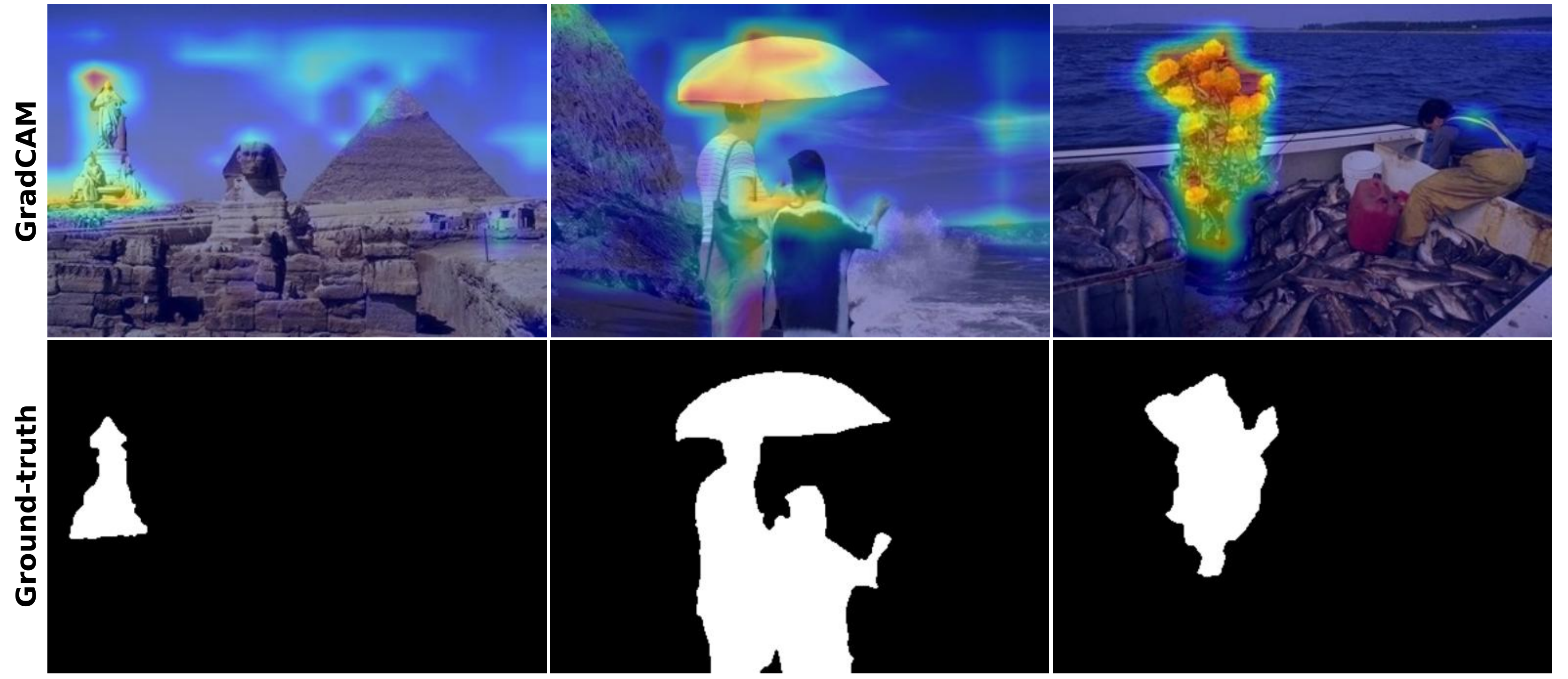}
    \vspace{-0.7cm}
    \small
    \caption{\small GradCAM visualizations of the backbone model after using ReVi.}
    \label{fig:cam}
\end{figure}

\begin{table*}[!t]
\centering
\small
\caption{\small Performance comparison between convolution and ReVi modules under equal parameter count.}
\vspace{-0.4cm}
\resizebox{\textwidth}{!}{
\begin{tabular}{llcccccccccc|cc}
\toprule
\multirow{2}{*}{Backbone} & \multirow{2}{*}{Adapter} & \multicolumn{2}{c}{CASIA1} & \multicolumn{2}{c}{Columbia} & \multicolumn{2}{c}{Coverage} & \multicolumn{2}{c}{NIST16} & \multicolumn{2}{c}{IMD20} & \multicolumn{2}{c}{Average} \\
\cmidrule(lr){3-4} \cmidrule(lr){5-6} \cmidrule(lr){7-8} \cmidrule(lr){9-10} \cmidrule(lr){11-12} \cmidrule(lr){13-14}
 & & AUC & F1 & AUC & F1 & AUC & F1 & AUC & F1 & AUC & F1 & AUC & F1 \\
\midrule
%+MOE       &  \\
\multirow{2}{*}{TinySAM \textit{w.}} & Convolution Modules & 64.84 & 29.68 & 70.75 & 43.31 & 55.90 & 11.49 & 65.18 & 26.67 & 57.97 & 14.48 & 62.93 & 25.13\\
& \textbf{ReVi} & 70.25 & 38.79 & 77.02 & 55.41 & 61.27 & 21.35 & 63.04 & 24.00 & 59.39 & 16.24 & 66.19 & 31.16\\
\midrule
\multirow{2}{*}{SegFormer \textit{w.}} & Convolution Modules & 66.29 & 31.13 & 79.70 & 59.30 & 58.24 & 16.00 & 64.79 & 26.70 & 63.03 & 20.92 & 66.41 & 30.81\\
& \textbf{ReVi}       & 82.03 & 42.73 & 83.62 & 60.25 & 72.31 & 28.93 & 75.94 & 28.73 & 78.49 & 26.20 & 78.48 & 37.37\\
\bottomrule
\end{tabular}}
\label{tab:Equal_Parameter}
\end{table*}

\begin{table*}[!th]
\centering
\small
\caption{\small Effect of adapter placement in our method (ReVi) on TinySAM.}
\vspace{-0.4cm}
%\resizebox{\textwidth}{!}{
\begin{tabular}{ccccccccccc|cc}
\toprule
\multirow{2}{*}{Placement} & \multicolumn{2}{c}{CASIA1} & \multicolumn{2}{c}{Columbia} & \multicolumn{2}{c}{Coverage} & \multicolumn{2}{c}{NIST16} & \multicolumn{2}{c}{IMD20} & \multicolumn{2}{c}{Average} \\
\cmidrule(lr){2-3} \cmidrule(lr){4-5} \cmidrule(lr){6-7} \cmidrule(lr){8-9} \cmidrule(lr){10-11} \cmidrule(lr){12-13}
& AUC & F1 & AUC & F1 & AUC & F1 & AUC & F1 & AUC & F1 & AUC & F1 \\
\midrule
\textit{En.} + \textit{De.}  & 66.74 & 31.03 & 79.54 & 58.93 & 63.47 & 23.31 & 62.17 & 20.04 & 60.09 & 16.55 & 66.40 & 29.97 \\
\textit{De.}        & 66.50 & 29.54 & 82.48 & 63.17 & 66.61 & 27.02 & 62.91 & 18.58 & 60.29 & 16.12 & 67.76 & 30.89 \\
\textit{En.}        & 70.25 & 38.79 & 77.02 & 55.41 & 61.27 & 21.35 & 63.04 & 24.00 & 59.39 & 16.24 & 66.19 & 31.16 \\
\bottomrule
\end{tabular}%}
\label{tab:placement}
\end{table*}

\begin{table*}[!t]
\centering
\small
\caption{\small Different learnable deviation $\bm{\Delta}$ in the MKE module of  our method (ReVi) on TinySAM.}
\vspace{-0.4cm}
%\resizebox{\textwidth}{!}{
\begin{tabular}{ccccccccccc|cc}
\toprule
Deviation & \multicolumn{2}{c}{CASIA1} & \multicolumn{2}{c}{Columbia} & \multicolumn{2}{c}{Coverage} & \multicolumn{2}{c}{NIST16} & \multicolumn{2}{c}{IMD20} & \multicolumn{2}{c}{Average} \\
\cmidrule(lr){2-3} \cmidrule(lr){4-5} \cmidrule(lr){6-7} \cmidrule(lr){8-9} \cmidrule(lr){10-11} \cmidrule(lr){12-13}
$\bm{\Delta}$ & AUC & F1 & AUC & F1 & AUC & F1 & AUC & F1 & AUC & F1 & AUC & F1 \\
\midrule
DCT     & 67.90 & 33.40 & 81.80 & 63.96 & 60.15 & 17.95 & 63.30 & 21.72 & 58.82 & 14.26 & 66.39 & 30.26\\
DCPM    & 66.63 & 31.26 & 76.12 & 52.25 & 59.86 & 17.26 & 62.86 & 20.89 & 58.14 & 13.47 & 64.72 & 27.03\\
Ours    & 70.25   & 38.79 & 77.02 & 55.41 & 61.27 & 21.35 & 63.04 & 24.00 & 59.39 & 16.24 & 66.19 & 31.16 \\
\bottomrule
\end{tabular}%}
\label{tab:delta}
\end{table*}

\subsection{Ablation Study}
In this section, we conduct a series of ablation studies on our method using two backbone models, TinySAM and SegFormer, under the Pre-trained protocol. Due to space limitations, ablation results for SegFormer and other settings are provided in the \textit{Supplementary}.

\smallskip\noindent\textbf{Comparison between Parameter-Equal Substitute and ReVi.}
To rigorously validate the effectiveness of our proposed ReVi, we replace it with a stack of convolutions while keeping the total number of parameters equal to that of ReVi. Specifically, we use five convolutional layers used in ReVi. As shown in Table~\ref{tab:Equal_Parameter}, under two different backbone networks, TinySAM and SegFormer, ReVi consistently outperforms the convolutional counterpart. These results demonstrate that the performance gains of our method stem from the design of ReVi itself, rather than simply from an increase in model parameters.

\smallskip\noindent\textbf{Where to Insert Adapters.} Note that all backbone models are based on Transformers, containing a number of blocks with an attention mechanism. In this part, we investigate the effect of adapter placement within the network. Specifically, we consider three settings: 1) inserting adapters into every encoder block; 2) inserting adapters into every decoder block; and 3) inserting adapters into every block of both the encoder and the decoder.
The results in Table~\ref{tab:placement} demonstrate that incorporating adapters solely in the encoder yields the best performance. 
We conjecture that the encoder plays a more important role in feature extraction. Fine-tuning this component is thereby more effective in guiding the backbone model to focus on manipulation traces. In contrast, the decoder primarily serves to reconstruct the prediction map from the encoded features. {As a result, adapting the decoder is less likely to refine feature representations, thus resulting in a performance drop.}

\smallskip\noindent\textbf{Learnable Deviation $\bm{\Delta}$.} In the main experiments, we use a single convolutional layer to approximate this deviation in the MKE module, formulated as $\Delta \approx \mathrm{Conv}(\bm{X}^k)$. In this part, we further investigate two alternative formulations: 1) \textit{Frequency analysis}: Frequency knowledge has been proven effective for forensics~\cite{HIFInet,objectformer}. Inspired by~\cite{10654876}, we first convert the feature $\bm{X}^k$ to frequency maps using Discrete Cosine Transform (DCT). Then we construct a selection map that is trained to adaptively select manipulation-related information; 2) \textit{Central difference analysis}: Extracting central difference could help expose manipulation traces~\cite{texture}. Therefore, we apply DCPM proposed in~\cite{wu2025rpcanet_pp} on $\bm{X}^k$. The results in Table~\ref{tab:delta} show that our single-layer convolution achieves the best performance.

\begin{table*}[thbp]
\centering
\small
\caption{\small The influence of semantic knowledge on TinySAM.}
\vspace{-0.4cm}
\resizebox{\textwidth}{!}{
\begin{tabular}{ccccccccccc|cc}
\toprule
\multirow{2}{*}{Setting} & \multicolumn{2}{c}{CASIA1} & \multicolumn{2}{c}{Columbia} & \multicolumn{2}{c}{Coverage} & \multicolumn{2}{c}{NIST16} & \multicolumn{2}{c}{IMD20} & \multicolumn{2}{c}{Average} \\
\cmidrule(lr){2-3} \cmidrule(lr){4-5} \cmidrule(lr){6-7} \cmidrule(lr){8-9} \cmidrule(lr){10-11} \cmidrule(lr){12-13}
& AUC & F1 & AUC & F1 & AUC & F1 & AUC & F1 & AUC & F1 & AUC & F1 \\
\midrule
(Frozen) TinySAM \textit{wo.} pretrained weights \textit{w.} \textbf{ReVi}    & 59.61 & 17.96 & 70.24 & 43.84 & 58.35 & 17.04 & 59.91 & 17.40 & 59.23 & 14.67 & 61.47 & 22.18\\
(Unfrozen) TinySAM \textit{wo.} pretrained weights \textit{w.} \textbf{ReVi}  & 65.48 & 30.90 & 76.78 & 55.21 & 57.98 & 14.87 & 65.31 & 26.08 & 60.08 & 17.44 & 65.13 & 28.90\\
TinySAM \textit{w.} \textbf{ReVi}      & 70.25 & 38.79 & 77.02 & 55.41 & 61.27 & 21.35 & 63.04 & 24.00 & 59.39 & 16.24 & 66.19 & 31.16\\
\bottomrule
\end{tabular}}
\label{tab:seminfluence}
\end{table*}

\subsection{Further Analysis.}

\smallskip\noindent\textbf{Effect of SKA and MKE Modules.}
Fig.~\ref{fig:revicam} illustrates the GradCAM maps attended by SKA and MKE modules using TinySAM on the CASIAv1 dataset. Specifically, we use the features from the first transformer block for illustration. As expected, SKA and MKE modules can guide the model to focus on the authentic background regions and the manipulated target regions, respectively.

\begin{figure}[!t]
    \centering
    \includegraphics[width=\linewidth]{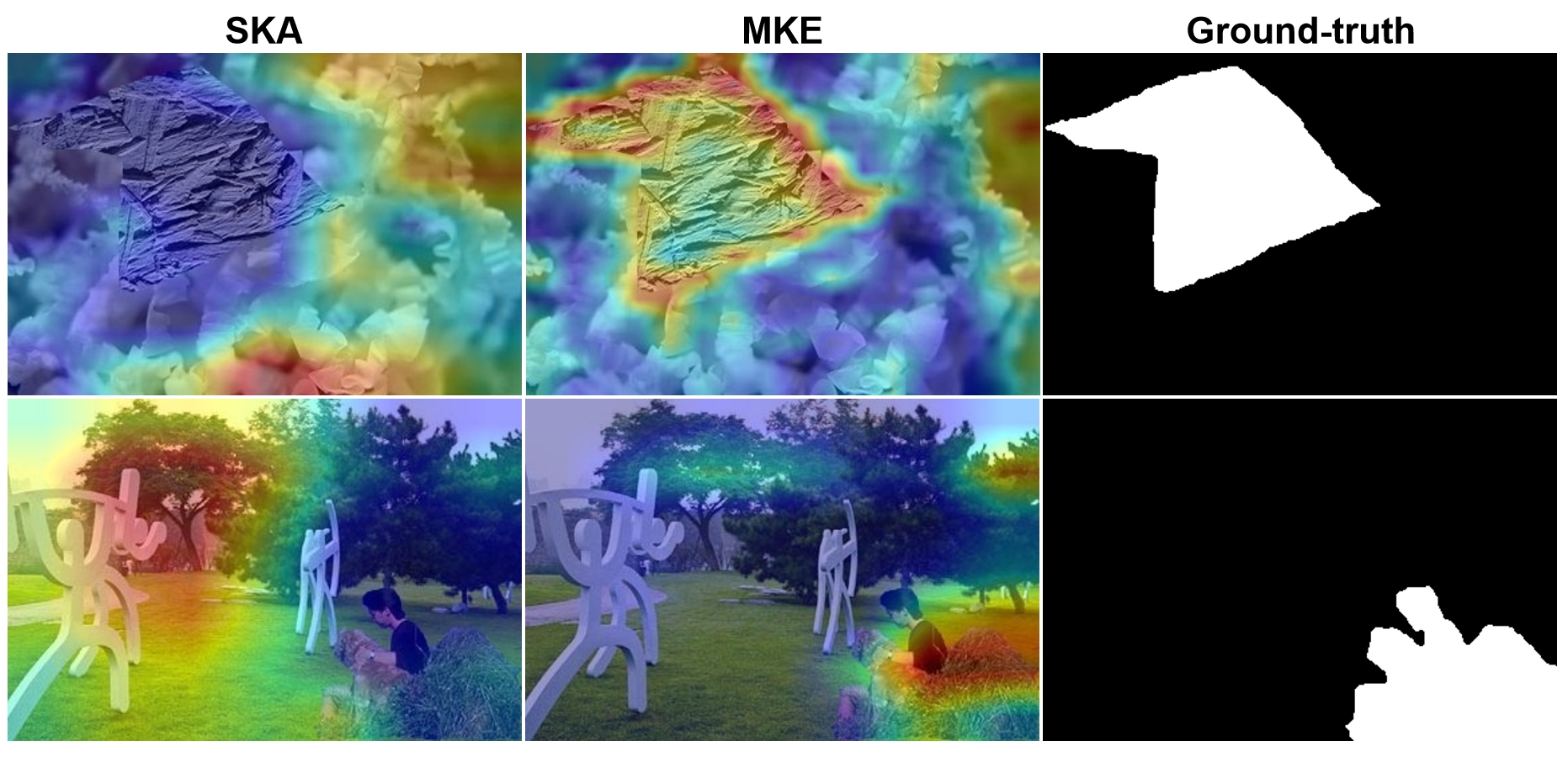}
    \vspace{-0.7cm}
    \small
    \caption{\small GradCAM visualizations of the SKA module and MKE module. }
    \label{fig:revicam}
    % \vspace{-0.4cm}
\end{figure}

\smallskip\noindent\textbf{The Influence of Semantic Priors.}
We conduct experiments to verify the influence of semantic priors in the vision models. Specifically, we study two settings: using TinySAM without its pretrained weights to remove semantic knowledge, and then either freezing or unfreezing its initial weights during adaptation with ReVi. As shown in Table~\ref{tab:seminfluence}, the performance of TinySAM without pretrained weights drops significantly. Even when all weights are made trainable by unfreezing TinySAM, its performance still fails to surpass that achieved with pretrained weights in the main experiments. These results indicate that leveraging semantic priors significantly benefits the IML task, and such knowledge is difficult to recover using only IML datasets.

\smallskip\noindent\textbf{Comparison of Different Fine-Tuning Methods.}
Building upon the preliminary experiments, we further study LDM and SegFormer under full fine-tuning, partial fine-tuning with LoRA versus our ReVi. Table~\ref{tab:fft-lora-revi} presents the comparison results on CASIAv1, NIST16, and IMD20 under Pre-trained protocol, where our ReVi substantially outperforms the full fine-tuning and LoRA baseline. We present the complete comparison results of the Pre-trained and Fine-tuned protocols in \textit{Supplementary}.

\begin{table}[!t]
    \centering
    \small
    \caption{\small Comparison results of different fine-tuning methods under Pre-trained protocol.}
    \vspace{-0.4cm}
    \resizebox{0.5\textwidth}{!}{
    \begin{tabular}{lcccccc}
    \toprule
         \multirow{2}{*}{Method} & \multicolumn{2}{c}{CASIAv1} & \multicolumn{2}{c}{NIST16} & \multicolumn{2}{c}{IMD20}\\
         \cmidrule(lr){2-3}  \cmidrule(lr){4-5}  \cmidrule(lr){6-7}
         & AUC & F1 & AUC & F1 & AUC & F1\\
         \midrule
        LDM \textit{w.} Full FT       & 65.73 & 29.75 & 61.57 & 19.34 & 58.55 & 13.87\\
        LDM \textit{w.} LoRA    & 67.84 & 33.17 & 60.44 & 16.22 & 58.43 & 15.91 \\
        LDM \textit{w.} \textbf{ReVi}               & 71.80 & 39.61 & 64.96 & 18.47 & 62.70 & 15.36\\
        \midrule
        SegFormer \textit{w.} Full FT   & 77.86 & 36.45 & 76.41 & 28.04 & 74.57 & 21.28\\
        SegFormer \textit{w.} LoRA         & 79.89 & 39.17 & 77.36 & 28.32 & 76.98 & 23.07\\
        SegFormer \textit{w.} \textbf{ReVi}         & 82.03 & 42.73 & 75.94 & 28.73 & 78.49 & 26.20\\
        % \midrule
        % Swin \textit{w.} Full FT  & 76.07 & 32.16 & 74.74 & 29.83 & 76.07 & 22.74\\
        % Swin \textit{w.} LoRA        & 78.20 & 36.42 & 77.02 & 29.32 & 76.20 & 23.14 \\
        % Swin \textit{w.} \textbf{ReVi}        & 81.07 & 38.80 & 78.29 & 30.74 & 76.92 & 24.36\\
         \bottomrule
    \end{tabular}}
    \label{tab:fft-lora-revi}
\end{table}

\smallskip\noindent\textbf{Complexity.}
Table~\ref{tab:param} exhibits the number of trainable parameters in the backbone models, our adapter, and their corresponding ratios.
The number of trainable parameters in our method scales with the size of the backbone model, but only a small number of parameters is required (approximately $13\%$), resulting in a reasonable computational overhead compared to dedicated methods. We additionally analyze the parameter counts of the compared methods. We observe that in the official code released by EVP~\cite{evp}, the backbone network is not frozen, which differs from the description in the original paper. To ensure a fair comparison, we adopt the setting used in their code. The results indicate that our method incurs only a minimal computational overhead, highlighting its potential for scalable deployment.

\begin{table}[!t]
    \centering
    \small
    \caption{\small Parameter analysis in millions (M). }
    \vspace{-0.4cm}
    %\resizebox{\textwidth}{!}{
    \begin{tabular}{lccc}
    \toprule
         Method
         & Frozen & Trainable & Rate \\
         \midrule
         PSCC-Net~\cite{pscc}   & 0 & 3.668 & 100.00$\%$\\ 
         TruFor~\cite{trufor}     & 0.557 & 66.562 & 99.17$\%$\\
         EVP~\cite{evp}        & 0 & 64.521 & 100.00$\%$\\
         IML-ViT~\cite{imlvit}    & 0 & 91.729 & 100.00$\%$ \\ 
         ACBG~\cite{ACBG}       & 0 & 51.798 & 100.00$\%$\\ 
         Sparse-ViT~\cite{sparsevit} & 0 & 50.344 & 100.00$\%$\\
         MPC~\cite{mpc}        & 0 & 45.035 & 100.00$\%$\\
         Mesorch~\cite{mesorch}    & 0 & 85.754 & 100.00$\%$ \\ 
         \midrule
         LDM \textit{w.} \textbf{ReVi} & 865.922 & 56.480 & 6.52$\%$\\ 
         TinySAM \textit{w.} \textbf{ReVi}  & 10.130 & 2.248 & 22.19$\%$\\
         SegFormer \textit{w.} \textbf{ReVi} & 13.678 & 1.931 & 14.12$\%$\\
         Swin Transformer \textit{w.} \textbf{ReVi} & 59.827 & 6.021 & 10.06$\%$ \\ 
         \bottomrule
    \end{tabular}%}
    \label{tab:param}
\end{table}

\begin{table}[!t]
  \centering
  \small
  \caption{\small Robustness performance of different methods.}
  \vspace{-0.4cm}
  \resizebox{0.5\textwidth}{!}{
  \begin{tabular}{lcc|c}
    \toprule
    \multirow{2}{*}{Distortion} & \multicolumn{3}{c}{Method} \\
    \cmidrule(l){2-4}
     & Sparse-ViT~\cite{sparsevit} & Mesorch~\cite{mesorch} & Ours \\
    \midrule
    No Distortion & 75.99 & 82.61 & \textbf{88.52} \\
    \midrule
    Blur ($k=3$) & 75.04 & 79.34 & \textbf{85.72}  \\
    Blur ($k=15$) & 69.53 & 62.57 & \textbf{85.65} \\
    \midrule
    Compress ($q=100$) & 75.77 & 82.38 & \textbf{84.30} \\
    Compress ($q=50$)  & 74.35 & 80.43 & \textbf{83.39} \\
    \midrule
    Resize ($0.78\times$) & 75.43 & 81.85 & \textbf{85.24} \\
    Resize ($0.25\times$) & 73.28 & 74.36 & \textbf{86.52}  \\
    \midrule
    Noise ($\sigma=3$) & 74.50 & 80.19 & \textbf{82.87}  \\
    Noise ($\sigma=15$) & 56.79 & 50.02 & \textbf{66.49} \\
    \bottomrule
  \end{tabular}}
  \label{tab:distortion_results}
\end{table}

\smallskip\noindent\textbf{Robustness.}
  To analyze the impact of varying degrees of post-processing on IML, we follow the setting of~\cite{SPAN,noiseguidance,pscc} and apply distortions to the images, including resizing the image to multiple scales, applying Gaussian blur with a kernel size of $k$, adding Gaussian noise with a standard deviation of $\sigma$, and performing JPEG compression with a quality factor of $q$. The experiment is conducted on the NIST16 dataset using TinySAM. The results shown in Table~\ref{tab:distortion_results} indicate that our method can resist distortions to some degree and performs better than other methods such as Sparse-ViT and Mesorch. 

\smallskip\noindent\textbf{Limitations.}
Since our method relies on off-the-shelf vision models, its performance is inherently associated with the capacity of these models. High-capacity models, which have learned rich semantic features, can fully leverage our method to capture subtle manipulation traces. However, when a very basic or low-capacity vision model is used, the effectiveness of our method may be compromised.

\section{Conclusions}
  This paper investigates the hypothesis that general semantic priors can enhance IML and introduces a novel adapter, ReVi, which can be seamlessly integrated into off-the-shelf general-purpose vision models to effectively repurpose them for IML tasks. Its design is inspired by the principles of Robust Principal Component Analysis, reformulating the decomposition into semantic-redundant and manipulation-specific components, and approximating this process through a newly devised deep architecture. Importantly, the proposed adapters are plug-and-play and show consistent effectiveness across a variety of off-the-shelf vision models, underscoring their potential for developing scalable IML frameworks.
{
    \small
    \bibliographystyle{ieeenat_fullname}
    \newpage
    \bibliography{main}
}

% WARNING: do not forget to delete the supplementary pages from your submission 
% \input{sec/X_suppl}

\end{document}